\documentclass[fleqn,10pt]{wlscirep}
\usepackage[utf8]{inputenc}
\usepackage[T1]{fontenc}
\usepackage[section]{placeins} 
\title{\LARGE Cross-Linguistic Analysis of Memory Load in Sentence Comprehension: Linear Distance and Structural Density}

\author[1,*]{Krishna Aggarwal}
\affil[1]{Department of Biological Sciences, Indian Institute of Science Education and Research (IISER), Mohali, India.}

\affil[*]{Corresponding author : aggarwalkrishna2811@gmail.com;\newline \hspace{1em}Alternate : ms21169@iisermohali.ac.in}

\begin{abstract}
This study examines whether sentence-level memory load in comprehension is better explained by linear proximity between syntactically related words or by the structural density of the intervening material. Building on locality-based accounts and cross-linguistic evidence for dependency length minimization, the work advances Intervener Complexity—the number of intervening heads between a head and its dependent—as a structurally grounded lens that refines linear distance measures. Using harmonized dependency
treebanks and a mixed-effects framework across multiple languages, the analysis jointly evaluates sentence length, dependency length, and Intervener Complexity as predictors of the Memory-load measure. Studies in Psycholinguistics have reported the contributions of feature interference and misbinding to memory load during processing. For this study, I operationalized sentence-level memory load as the linear sum of feature misbinding and feature interference for tractability; current evidence does not establish that their cognitive contributions combine additively. All three factors are positively associated with memory load, with sentence length exerting the broadest influence and Intervener Complexity offering explanatory power beyond linear distance. Conceptually, the findings reconcile linear and hierarchical perspectives on locality by treating dependency length as an important surface signature while identifying intervening heads as a more proximate indicator of integration and maintenance demands. Methodologically, the study illustrates how UD-based graph measures and cross-linguistic mixed-effects modelling can disentangle linear and structural contributions to processing efficiency, providing a principled path for evaluating competing theories of memory load in sentence comprehension.
\end{abstract}

\begin{document}
\flushbottom

\maketitle
\vspace{-1.5em}
\vspace{-0.6em}
%

\section*{Introduction}

Explaining cross-linguistic preferences in word order requires theories that link grammatical organization to real-time processing under memory constraints, an approach that traces back to parsing-based accounts of universals and locality in syntax \cite{hawkins1990}. In this tradition, dependency-based efficiency has provided a unifying quantitative signature: languages tend to favour word orders in which syntactically related words (heads and dependents) are kept close, thereby reducing integration and maintenance costs during incremental comprehension and production \cite{temperley2018}. The core empirical generalization, known as Dependency Length Minimisation (DLM), measures the linear distance between heads and dependents, and has been widely used to capture processing difficulty and memory burden in corpus studies and typological comparisons \cite{liu2017}. Large-scale cross-linguistic analyses have shown that observed sentences in many languages exhibit substantially shorter total dependency length than carefully matched randomized baselines, indicating a systematic bias toward locality that extends across sentence lengths and constructions \cite{futrell2015}. Together with the theory that formalizes typed combinatory operations and logical syntax, these findings situate locality as a product of resource-bounded derivational processes that structure how dependencies are created and interpreted \cite{morrill2011}.

At the same time, linear distance is best viewed as a proxy for deeper structural work performed by the parser. Reviews emphasise that different linguistic systems implement locality pressures via distinct means (word order, morphology, prosody), and that estimates of ``distance'' depend on how intervening structure is quantified \cite{temperley2018, liu2017}. A complementary perspective, therefore, foregrounds the \emph{complexity of the material that intervenes} between heads and dependents. Rather than counting intervening words, this view evaluates the number (and configuration) of intervening structural units---in particular, intervening syntactic heads---as the more direct index of integration and maintenance demands during parsing. we refer to this measure as \emph{Intervener Complexity}. Conceptually, each intervening head increases the number of commitments that must be sustained and the opportunities for similarity-based interference during retrieval, aligning the metric more closely with how typed derivations consume resources in categorial frameworks and how incremental parsers construct structure \cite{morrill2011, hawkins1990}.

Intervener Complexity thus complements Dependency Length in two ways. First, it distinguishes spans that are equally long linearly but differ in how much structure the parser must build and maintain: a stretch containing multiple heads can impose higher memory and interference costs than a stretch of similar length with fewer heads \cite{liu2017, temperley2018}. Second, it connects naturally to formal derivational burdens. In categorial grammar, more intervening heads typically entail additional combinatory steps and intermediate categories that must be held until discharge, increasing the risk of stack growth and retrieval competition \cite{morrill2011}. On this view, Intervener Complexity provides a structurally grounded lens on memory load that refines the linear locality captured by DLM, rather than replacing it.

Methodological advances now make it feasible to test these ideas at scale. Universal Dependencies (UD) offers harmonised, dependency-annotated corpora for many languages, enabling cross-linguistic measurement of both linear and structural locality with consistent annotation and facilitating baseline construction that preserves key tree properties \cite{nivre2018}. Leveraging UD, corpus studies can estimate the unique and joint contributions of Dependency Length, Intervener Complexity, and Sentence Length to processing-related outcomes, while accounting for between-language heterogeneity via mixed-effects modelling \cite{futrell2015, nivre2018}. This empirical setting aligns with the efficiency perspective summarised in recent reviews: locality effects should be detectable across typology, but their strength and interaction with other grammatical dimensions (argument structure, head-directionality, case marking) may vary, reflecting language-specific solutions to general processing pressures \cite{temperley2018, hawkins1990}.

The present study adopts Intervener Complexity as the primary explanatory lens on sentence-level memory load, while jointly modelling the contributions of Dependency Length and Sentence Length. Memory load is operationalised as a composite sentence-level measure sensitive to interference and misbinding risks, and predictors are quantified from UD-style dependency graphs. A linear mixed-effects model with language as a random intercept separates universal sentence-level effects from cross-linguistic baselines. The empirical results demonstrate three core findings. First, all predictors show positive and statistically significant associations with memory load, consistent with efficiency expectations: longer distances, more intervening heads, and longer sentences each correlate with higher processing demand. Second, \emph{Intervener Complexity exerts a reliably larger effect than Dependency Length}, indicating that structural density in the span between head and dependent contributes over and above linear separation, as predicted by derivational and memory-based accounts \cite{morrill2011, liu2017, temperley2018}. Third, \emph{Sentence Length is the dominant predictor} by a large margin, capturing global increases in representational load as sentences grow. At the same time, the substantial random-intercept variance for language confirms meaningful cross-linguistic baselines that are orthogonal to the fixed-effect hierarchy \cite{nivre2018}.

These outcomes reconcile two strands of the literature. On the one hand, they align with the cross-linguistic DLM evidence: linear distance matters and contributes positively to memory load \cite{futrell2015}. On the other hand, they vindicate the structural view that not all spans are equal: intervening heads---as loci of structure-building---are a better proximal determinant of integration and maintenance cost than word count alone, thereby explaining why some long dependencies are tolerated when structurally sparse,'' while some shorter spans are costly when structurally dense'' \cite{temperley2018, liu2017, morrill2011}. By placing Intervener Complexity at the centre and situating DLM within a broader structural-efficiency model, the study clarifies how linear and hierarchical factors jointly shape memory load, and why typological differences can arise despite shared efficiency pressures \cite{hawkins1990, temperley2018}. The availability of UD resources ensures that these inferences rest on consistent, multilingual evidence \cite{nivre2018}.

In sum, the contribution is twofold. Substantively, it provides cross-linguistic evidence that \emph{Intervener Complexity is a stronger predictor of sentence-level memory load than linear Dependency Length}, while confirming the dominant role of Sentence Length and the importance of language-level baselines. Methodologically, it demonstrates how dependency-graph measures and mixed-effects modelling over UD corpora can adjudicate between linear and structural notions of locality, advancing efficiency-based explanations of word order that are compatible with formal theories of composition and derivation \cite{morrill2011, hawkins1990, temperley2018, nivre2018, futrell2015, liu2017}.

\section*{Methods}

\subsection*{Dataset}
The study utilised the deep universal dependencies dataset \cite{deepud28}. The distributed dataset (\texttt{deep-ud-2.8-data.tgz}) contains CoNLL-U formatted files across multiple languages. These files were programmatically extracted and organized by language for downstream processing.
\subsection{Data Preprocessing}
Custom Python scripts (implemented in \texttt{data\_extract.py}) extracted raw sentences from the \texttt{"text"} metadata in each CoNLL-U file. A total of 23 typologically diverse languages were selected, and for each language, 500 sentences were randomly sampled to construct the final dataset comprising 11,500 sentences. The extraction process included file format validation, UTF-8 encoding checks, and ensured consistent sentence representation. Parallel processing using Python’s \texttt{ThreadPoolExecutor} accelerated the processing across languages. Sentences were stored in a nested dictionary keyed by language and later converted to a fully expanded and shuffled Pandas DataFrame before saving in CSV format (\texttt{Project\_data.csv}).

\subsection*{Feature Computation}

Feature extraction was performed via methods in \texttt{features\_class.py}. Dependency parsing leveraged \textbf{spaCy} with language-specific models listed in \texttt{models.txt}. For tokenization in Japanese, Chinese, and Korean, language-specific tokenizers—Janome, Jieba, and KoNLPy Okt, respectively—were employed. Dependency structures were represented as directed graphs using \textbf{NetworkX}, from which four linguistic measures were computed for each sentence:

\begin{itemize}
    \item \textbf{Memory Load}: I have considered Memory Load to be a composite measure defined as the sum of:
    \begin{enumerate}
        \item \textit{Feature Interference}: approximated by the extent of repeated dependency labels and part-of-speech tags within a sentence, which serve as a computational proxy for similarity-based interference (competition among items with overlapping features during sentence processing, where overlapping features cause competition.   \cite{Lewis2005, VanDyke2012}
        
        \item \textit{Feature Misbinding}: approximated by counting cases where nominal dependents (e.g., \texttt{nsubj}, \texttt{dobj}, \texttt{iobj}, \texttt{pobj}) are attached to non-root or non-clausal heads, which we treat as proxies for feature-binding failures in feature-to-head binding during syntactic processing.
        \cite{Dempsey2022, Villata2018Agreement}
    \end{enumerate}

Both feature interference and feature misbinding have been reported in the psycholinguistics literature as contributors to memory-related difficulty during sentence comprehension, since interference among overlapping features and binding failures each increase retrieval demands and processing cost \cite{Lewis2005, VanDyke2012, Villata2018German, Villata2018Agreement, Dempsey2022}. In this study, I operationalize memory load as a composite measure defined by the linear sum of these two components, providing a computationally tractable proxy for sentence-level working memory demands.

    \item \textbf{Dependency Length}: The sum of the absolute linear distances between syntactic heads and their dependents derived from adjacency matrices of dependency graphs.

    \item \textbf{Intervener Complexity}: A measure based on counting intervening nodes between head-dependent pairs within the dependency graph, providing an index of syntactic complexity.

    \item \textbf{Sentence Length}: Determined as the total number of tokens, counted via whitespace tokenization for alphabetic languages, and by Janome, Jieba, or KoNLPy Okt tokenizers for Japanese, Chinese, and Korean, respectively.
\end{itemize}

\subsection*{Data Aggregation}

The extracted features for all sentences and languages were aggregated into a nested dictionary (per language) within \texttt{data\_extract.py}. This dictionary was converted into a Pandas DataFrame, with sentence-level expansion and random shuffling applied to mitigate order effects. The resulting structured dataset was saved as \texttt{Project\_data.csv} for subsequent statistical modelling.

\subsection*{Statistical Analysis}

Statistical evaluation employed a linear mixed-effects model to investigate how syntactic features relate to memory load. The analysis was conducted in the \texttt{main.ipynb} notebook, using the Python \texttt{statsmodels} library. The model treated \textit{memory load} as the dependent variable, with \textit{dependency length}, \textit{intervener complexity}, and \textit{sentence length} as fixed effects predictors. To account for variability arising from linguistic differences, \textit{language} was modelled as a random intercept effect:

\[
\text{Memory Load} \sim \text{Dependency Length} + \text{Intervener Complexity} + \text{Sentence Length} + (1 \mid \text{Language})
\]

This approach allowed for robust inference on the fixed effects while controlling for intra-language correlations.

\section*{Results}
Sentence length has the most significant positive effect on Memory load, with an estimated coefficient of approximately 0.389 and an extremely large z-statistic (~62.39), indicating a strong and dominant influence among predictors.

\subsection*{Fixed-effect results}
\begin{figure}[h!] 
\centering
\includegraphics[scale=0.7]{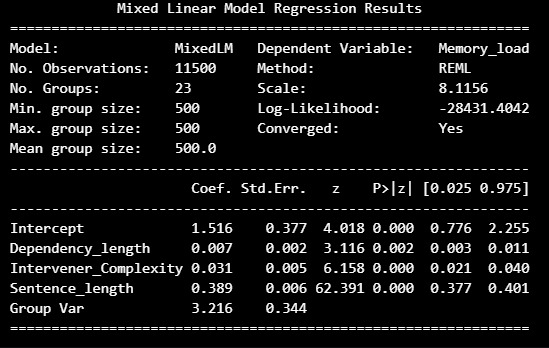} 
\caption{Mixed-effects model showing positive effects of sentence-level memory load predictors}
\label{fig:example-figure} 
\end{figure}

The REML linear mixed-effects model indicates that all three sentence-level predictors are positively and significantly associated with Memory load, with effect sizes spanning small to large magnitudes and accompanied by narrow 95\% confidence intervals that signal high estimation precision. Specifically, Dependency length shows a modest coefficient of approximately 0.007 with a z-statistic near 3.12, indicating a reliable but comparatively small contribution to memory demands; Intervener Complexity yields a moderate coefficient of about 0.031 with a z-statistic around 6.16, suggesting a more evident and more decisive influence than linear distance alone; and Sentence length exhibits a significant coefficient of roughly 0.389 with an exceptionally high z-statistic near 62.39, establishing it as the dominant predictor in both magnitude and statistical certainty. The relative ordering of effects—Sentence length overwhelmingly first, followed by Intervener Complexity, then Dependency length—aligns with the fixed-effects point estimates and their confidence intervals in the associated plots, reinforcing the conclusion that overall sentence size is the primary determinant of sentence-level memory load, while structural complexity and head–dependent distance provide additional, smaller positive contributions. Beyond fixed effects, the model estimates a substantial random-intercept variance for language (Group Var $ \approx 3.216$), which captures baseline cross-linguistic heterogeneity in memory load and justifies modelling language-specific shifts; critically, accounting for these random effects improves fit without altering the core hierarchy among fixed predictors, thereby separating universal predictors of memory demand from language-dependent baselines in a principled way.

\subsection*{Marginal patterns}

\begin{figure}[!htbp]
\centering
\begin{minipage}{0.32\textwidth}
  \centering
  \includegraphics[width=\textwidth]{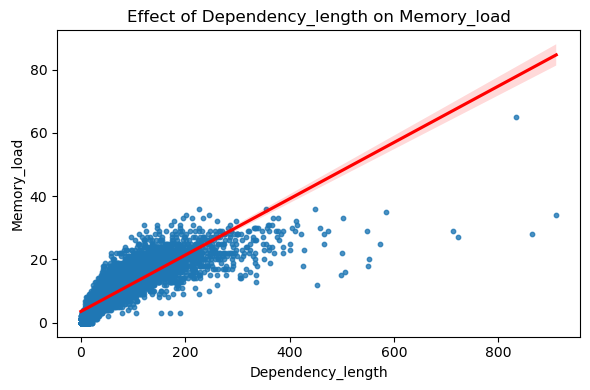} 
\end{minipage}\hfill
\begin{minipage}{0.32\textwidth}
  \centering
  \includegraphics[width=\textwidth]{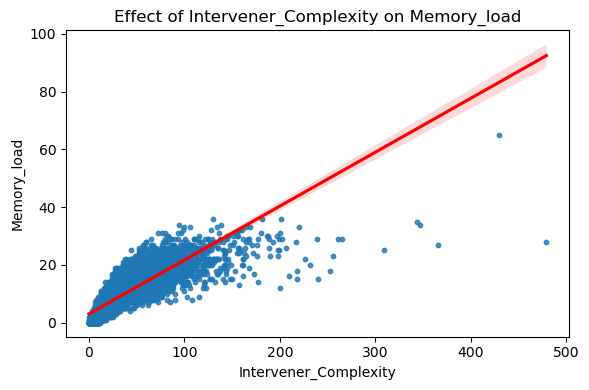} 
\end{minipage}\hfill
\begin{minipage}{0.32\textwidth}
  \centering
  \includegraphics[width=\textwidth]{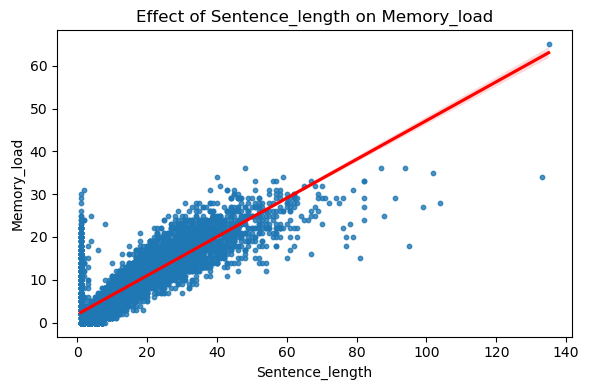} 
\end{minipage}
\caption{Positive effects of sentence-level predictors (Sentence length, Intervener complexity, Dependency length) on Memory load}
\label{fig:example-figure}
\end{figure}

Across the marginal-effect panels, each predictor shows a positive association with Memory load, but the slopes differ markedly in magnitude and precision. Dependency length displays a shallow, positive trend, indicating that increases in head–dependent distance are associated with only slight increases in memory demands; this aligns with the mixed-effects estimate of roughly 0.007 and its narrow confidence interval, which implies statistical reliability despite a modest effect size. Intervener Complexity presents a noticeably steeper positive slope, consistent with its larger coefficient of about 0.031 and a confidence interval that excludes zero; this pattern suggests that sentences with more intervening material impose a meaningfully greater memory burden than distance alone would predict. In contrast, Sentence length exhibits the steepest slope with tight confidence bands, mirroring its dominant fixed-effect estimate near 0.389 and very large z-statistic; this indicates that overall sentence size is the strongest and most precisely estimated driver of memory load in the observed data.

\subsection*{Cross-linguistic distribution}
\FloatBarrier
The violin plots across 23 languages reveal substantial within-language dispersion and apparent between-language heterogeneity in Memory load, with most distributions concentrated around 5–15 and extended right tails indicating occasional high-demand sentences. Languages such as English and Japanese exhibit broader spreads and higher outliers, whereas Norwegian and Korean display more compact distributions, consistent with sizable cross-linguistic baseline differences captured by the random intercept (variance $ \approx 3.216$). Modelling language as a random effect appropriately absorbs these baseline shifts and improves overall fit while preserving the fixed-effects ordering of predictors established by the mixed-effects estimates.

\begin{figure}[!htbp]
  \centering
  \includegraphics[width=19cm, height=13cm]{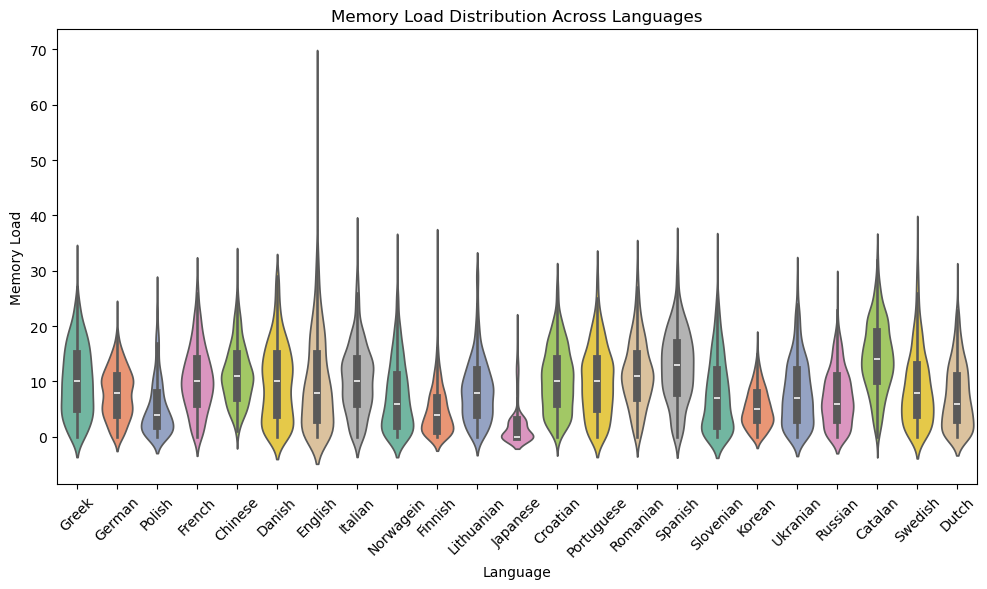}
  \caption{Violin plot showing the distribution of memory load across 23 languages.}
  \label{fig:violin}
\end{figure}
\FloatBarrier

\subsection*{Model fit and predictive accuracy}
\FloatBarrier
The mixed-effects model provided a strong account of sentence-level variability in Memory load and generalised well across languages. At the observation level, the model explained a substantial proportion of variance, with R\textsuperscript{2}$ \approx 80.91$\%; by this magnitude, the jointly specified fixed effects together with language as a random grouping factor captured the majority of systematic variation in memory load, indicating that the model structure isolates the dominant signal in the data rather than residual noise. Error magnitudes were modest in squared and absolute terms (MSE $ \approx 8.0970$; MAE $ \approx 2.0867$), implying close correspondence between predictions and observed values and suggesting that typical prediction errors are minor relative to the empirical range of Memory load. Complementing these global indices, comparisons of language-wise means for actual versus predicted Memory load showed near overlap for most languages, with only minor deviations in a small number of cases; this pattern is consistent with effective cross-linguistic generalisation and indicates that the model does not rely on idiosyncratic properties of any single language to achieve its overall fit. Methodologically, treating language as a random intercept absorbs baseline differences across languages while preserving the ordering and interpretability of the fixed effects, thereby enabling sentence-level predictors to be evaluated on a standard scale and ensuring that cross-language heterogeneity does not confound their estimated associations with Memory load. Taken together, these results support the adequacy of the fixed-effects specification, confirm the utility of the language-level random intercept in capturing cross-linguistic baselines, and demonstrate accurate, stable predictions at both sentence and language aggregation levels.

\begin{figure}[!htbp]
  \centering
  \includegraphics[width=\linewidth]{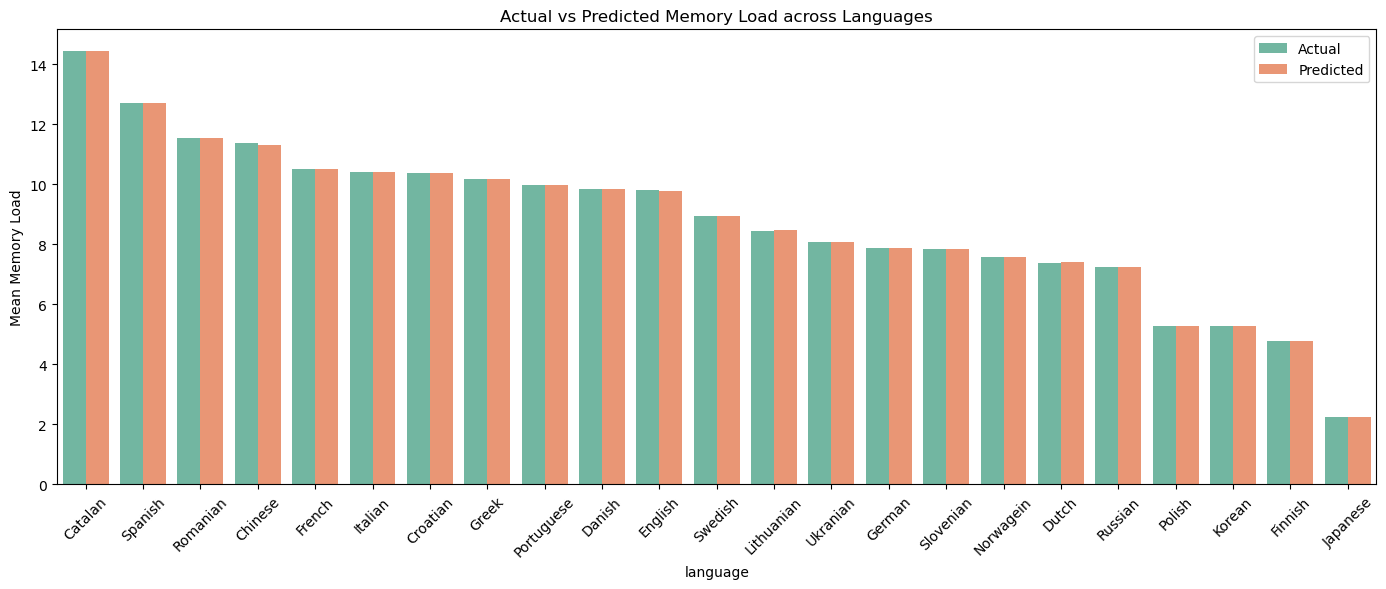}
  \caption{Observed vs. predicted mean memory load by language, showing close alignment (R² $ \approx 80.9$\% )with low error, indicating strong model fit.}
  \label{fig:violin}
\end{figure}
\FloatBarrier

\subsection*{Interpretation and implications}

Sentence length emerges as the principal determinant of working‑memory demand in this corpus, with its effect substantially exceeding those of linear distance and intervening material, indicating that overall sentence size most strongly drives increases in Memory load. Intervener Complexity exerts a reliably larger influence than dependency distance, consistent with the idea that intervening syntactic heads impose additional representational and retrieval burdens beyond mere head–dependent separation. At the same time, sizeable language-level variability underscores genuine cross‑linguistic differences in baseline memory demands; modelling language as a random intercept captures these shifts while preserving the fixed‑effects ordering, allowing generalizable inferences about sentence‑level predictors across languages.

\section*{Discussion}

This study examined how linear and structural locality relate to sentence-level memory load, asking whether structural density between heads and dependents adds explanatory value beyond linear distance. Across a typologically diverse sample, three predictors—Sentence Length, Intervener Complexity, and Dependency Length—were positively associated with the memory-load index. Two patterns stand out. First, Sentence Length had the most decisive influence, consistent with the idea that overall representational size constrains processing resources at the sentence level. Second, Intervener Complexity contributed more than Dependency Length, indicating that the number of intervening heads—sites of structure-building—captures integration and maintenance demands that word-count distance only partially reflects.
These findings suggest that linear and structural factors shape memory load in sentence comprehension. Linear proximity is important, but structural density is a more direct predictor when both are considered together. The observed ordering of effects—Sentence Length $>$ Intervener Complexity $>$ Dependency Length—offers a reconciliation: surface distance provides valuable information, yet the main burden lies in how many structural commitments must be maintained. Viewed typologically, the results align with the idea that languages implement locality through different mechanisms (such as word order, case marking, or prosody) and may differ in how linear and structural signals trade off. The language-level random effects confirm meaningful cross-linguistic baselines independent of the fixed-effect hierarchy, which is expected given variation in head-directionality, morphological richness, and construction types. Future work could explore whether the relative weights of linear and structural predictors correlate with typological features such as argument marking or head-final ordering. Methodologically, the analysis shows how harmonized dependency representations and mixed-effects models can disentangle linear and hierarchical contributions to locality. Intervener Complexity, introduced here as a structural predictor, formalizes a simple intuition: equal linear spans can differ in processing cost if one path crosses more heads. This measure complements traditional dependency length without replacing it and offers a practical tool for developing structure-preserving baselines in future cross-linguistic studies.

\subsection*{Limitations}

First, the way I measured memory load is a simplified choice: it is defined as the sum of feature misbinding and feature interference. Both come from fundamental theories of processing difficulty, but I am not claiming that the brain adds them together. The results should be read with this simplification in mind. Second, because the study is based on corpus data, it's possible to show associations between variables but cannot prove direct cause-and-effect. Third, although I tried to standardize the datasets, tokenizers, and parsers across languages, the analysis still inherits some known limitations — for example, differences in treebank coverage, imbalances in text genres, and parsing errors. Finally, our decision to sample a fixed number of sentences per language may influence the variability of the results. Using larger and more balanced datasets in the future will help make cross-linguistic comparisons more stable.

\section*{Data and Code Availability}
To ensure reproducibility, all code for sentence extraction, feature computation, and data processing has been made publicly available at:
\href{https://github.com/KrishnaAggarwal2003/Computational-Analysis-of-Memory-Load-in-Language-Comprehension}{https://github.com/KrishnaAggarwal2003/Computational-Analysis-of-Memory-Load-in-Language-Comprehension}

\end{document}